\documentclass[runningheads]{llncs}

\usepackage{graphicx}
\usepackage{comment}
\usepackage{amsmath,amssymb}
\usepackage{color}
\usepackage{url}
\usepackage{hyperref}

\newif\ifreview
\reviewfalse

\ifreview
	\usepackage{lineno}

	\linenumbers
\fi

\begin{document}



\def\GCPRTrack{Main Track}

\title{Robust-DefReg: A Robust Deformable Point Cloud Registration Method based on Graph Convolutional Neural Networks}

\ifreview
\else
	\titlerunning{Robust-DefReg: A Robust Deformable Point Cloud Registration Method}

	\author{Sara Monji-Azad\inst{1}\orcidID{0000-0002-2742-9961} \and
	Marvin Kinz\inst{1}\orcidID{0009-0008-8226-5905} \and
	J\"urgen Hesser\inst{1,2,3,4}\orcidID{0000-0002-4001-1164}}
	
	\authorrunning{S. Monji-Azad et al.}
	
	\institute{Mannheim Institute for Intelligent Systems in Medicine (MIISM), Medical Faculty Mannheim, Heidelberg University, Mannheim, Germany
	\email{\{sara.monjiazad,juergen.hesser\}@medma.uni-heidelberg.de} \\
    \email{\{m.kinz\}@stud.uni-heidelberg.de} \\
	\and Interdisciplinary Center for Scientific Computing (IWR), Heidelberg University, Heidelberg, Germany\\
        \and Central Institute for Computer Engineering (ZITI), Heidelberg University, Heidelberg, Germany \\
        \and CZS Heidelberg Center for Model-Based AI, Heidelberg University, Mannheim, Germany}

\fi

\maketitle              

\begin{abstract}

Point cloud registration is a fundamental problem in computer vision that aims to estimate the transformation between corresponding sets of points. Non-rigid registration, in particular, involves addressing challenges including various levels of deformation, noise, outliers, and data incompleteness. This paper introduces Robust-DefReg, a robust non-rigid point cloud registration method based on graph convolutional networks (GCNNs). Robust-DefReg is a coarse-to-fine registration approach within an end-to-end pipeline, leveraging the advantages of both coarse and fine methods. The method learns global features to find correspondences between source and target point clouds, to enable appropriate initial alignment, and subsequently fine registration. The simultaneous achievement of high accuracy and robustness across all challenges is reported less frequently in existing studies, making it a key objective of the Robust-DefReg method. The proposed method achieves high accuracy in large deformations while maintaining computational efficiency. This method possesses three primary attributes: high accuracy, robustness to different challenges, and computational efficiency. The experimental results show that the proposed Robust-DefReg holds significant potential as a foundational architecture for future investigations in non-rigid point cloud registration. The source code of Robust-DefReg is available.\par

\end{abstract}
\section{Introduction}
Point cloud registration is a fundamental area within computer vision that aims to estimate transformations between corresponding point sets \cite{monji2023review}. It can be found in various applications, including virtual/augmented reality \cite{mahmood20193d}, LiDAR applications \cite{wang2018lidar}, quality control in industrial processes \cite{wang2019applications}, etc. The goal of point cloud registration is to minimize the error between the transformed point cloud and the target one \cite{deng2022survey}. Registration approaches can be divided into rigid and non-rigid transformations. Specifically, rigid registration refers to situations where the object's geometric properties and shape remain unchanged under affine transformations like translation and rotation. On the other hand, non-rigid registration involves determining the deformation field for the source point cloud \cite{monji2023review}. Finding stable corresponding points and a proper transformation function that can transfer target points to source ones are the main challenges for a registration method \cite{Castellani.2020}. Additionally, being robust to various levels of deformation, noise, outliers, and data incompleteness are also significant challenges for a registration method  \cite{wang2022gp}.

The recent state of the art regarding non-rigid point cloud registration can be classified from different points of view. In one aspect, the existing approaches can be categorized into coarse and fine approaches. In another categorization, the topic is classified into feature-based methods that focus on matching distinctive features, while the other methods directly model and estimate the deformation field. Additionally, learning-based methods and non-learning-based methods can be considered as a further categorization \cite{monji2023review}. \par

In this paper, we propose a coarse-to-fine method called Robust-DefReg, which is based on feature learning. Our method is inspired by \cite{hansen2021deep} and presents a feature descriptor network with high accuracy, robustness to different challenges, and computational efficiency. To achieve these three goals, Robust-DefReg is an end-to-end pipeline that takes advantage of Graph Convolutional Networks (GCNNs) \cite{zhang2019graph}, Loopy Belief Propagation (LBP), and spatial transformer networks (T-Net) \cite{jaderberg2015spatial}. Additionally, in \cite{hansen2021deep}, the method focuses on registering key points of deformed lungs. Robust-DefReg aims to enhance its capabilities to serve as a general registration technique for various types of point clouds, not only medical data.

The complexity of non-linear deformations and the difficulty in accurately modeling and capturing deformations at different levels led us to propose a coarse-to-fine registration method. To this end, Robust-DefReg starts from an initial guess and then enhances it in the next steps. Although finding an appropriate initial alignment is a good strategy, it is also challenging, especially when dealing with shape differences or missing data. Robust-DefReg tackles the difficulty of non-linear deformations by leveraging GCNNs to learn global features. This initial alignment serves as a foundation for subsequent feature layers, improving the registration process. Furthermore, in the proposed feature descriptor network, T-Net is used to find the best linear transformation parameters, making the feature descriptor network robust to different transformations.

From another perspective, robust techniques are necessary to handle outliers and missing data, as well as to avoid propagating errors during refinement. Moreover, the computational complexity of these methods is high, especially for large-scale point clouds \cite{yu2021cofinet}. In our proposed method, although the size of the feature descriptor network becomes larger, the computational efficiency is still as good as similar methods. The combination of T-Net, graph network, and LBP makes the proposed method robust to different challenges, resulting in high registration accuracy and acceptable computational efficiency.

By integrating robust mathematical models, efficient algorithms, and careful parameter selection, Robust-DefReg presents the following statements:

\begin{itemize}
    \item An end-to-end method for non-rigid point cloud registration using Graph Convolutional Neural Networks (GCNN), T-Net, and Loopy Belief Propagation (LBP).
    
    \item A coarse-to-fine method that improves accuracy without increasing computation time.

    \item Evaluation of the proposed method's robustness to challenges such as noise, outliers, and data incompleteness, demonstrating higher accuracy while maintaining computational efficiency.

    \item Achievement of higher accuracy compared to state-of-the-art methods, particularly in scenarios involving large deformations.
\end{itemize}

A supplementary paper has been published alongside the current paper, providing detailed information on network architectures and explaining the dataset preparation process with various challenges.\par

\section{Related Work}
Certain learning architectures, including convolutional neural networks (CNN), recurrent neural networks (RNN), graph convolutional neural networks (GCNN), and multilayer perceptron (MLP), are commonly employed in learning-based point cloud registration techniques \cite{monji2023review}. This section provides an outline of non-rigid registration methods that utilize learning approaches based on GCNN. Subsequently, to elaborate on the advantages of the proposed method, an overview of coarse and fine feature-based methods, as well as existing methods focusing on robustness, will be presented.\par 

\textbf{GCNNs.} 
It is a type of neural network designed specifically for processing data represented as graphs or networks \cite{zhang2019graph}. They take into account the graph structure and neighboring nodes in a convolutional manner, enabling the learning of geometric features from point clouds with neighborhood relations. One notable application of GCNN in the medical field is presented by \cite{hansen2021deep}. They combine edge convolutions to capture geometric features and use differentiable LBP to regularize displacements on a k-nearest neighbor graph (k-NNG), enabling 3D lung registration \cite{hansen2021deep}. The GCNN structure used in their study is inspired by the EdgeConv network architecture \cite{wang2019dynamic}, which effectively represents the relationship between points and their neighbors. Another study by \cite{hansen2019learning} proposes a dynamic GCNN for point cloud registration. This approach involves learning robust correspondences between source and target point sets using an EdgeConv-like architecture, followed by probabilistic refinement using the learned features in the coherent point drift (CPD) algorithm. SyNoRiM \cite{huang2022multiway} is another registration approach that utilizes a CNN with a point cloud graph as input. It assumes a fully connected graph, where the vertices represent input point clouds and edges represent graph connectivity. SyNoRiM computes pairwise correspondences parameterized using functional maps and tackles the occlusion problem by learning non-orthogonal basis functions for deformations. However, it still faces challenges in scaling to large scenes. \par

\textbf{Coarse vs. Fine Feature-based Methods.}
Coarse feature-based methods aim to quickly align point clouds by identifying and matching distinctive features, such as key points or landmarks \cite{wang2019prnet}. These methods typically use efficient algorithms or heuristics to estimate an initial transformation \cite{yu2021cofinet}. Although coarse feature-based methods provide rapid alignment, they may lack precision and struggle with capturing fine details in the point clouds, as shown in studies such as \cite{deng2018ppfnet}, \cite{deng2018ppf}, and \cite{gojcic2019perfect}. On the other hand, fine feature-based methods focus on refining the initial alignment by iteratively optimizing the transformation to minimize the misalignment between corresponding features. These methods are computationally more intensive and may require multiple iterations, but they can handle complex deformations and noise better, resulting in a more precise alignment. Studies such as DRC-Net \cite{li2020dual}, Patch2Pix \cite{zhou2021patch2pix}, and LoFTR \cite{sun2021loftr} utilize a method that gradually refines a coarse-to-fine mechanism to address the inherent issue of repeatability in keypoint detection. This approach is employed to enhance the overall performance of the detection process. \par

\textbf{Robustness.}
As discussed in \cite{monji2023review}, another point of reference to study the existing approaches focuses on their robustness to various mentioned challenges, such as different deformation levels \cite{Li.2020b,Yang.2018b}, noise \cite{ge2014non,ge2015non}, outlier levels \cite{shimada2019dispvoxnets,myronenko2010point}, as well as data incompleteness \cite{Bernreiter.2021}. However, a key gap in the existing literature lies in the lack of methods that simultaneously achieve high accuracy across all the aforementioned challenges. In this context, our proposed method represents a significant step towards tackling these challenges.\par

\section{Proposed Method}
\begin{figure}[!ht]
\centering
\includegraphics[width=12.15cm]{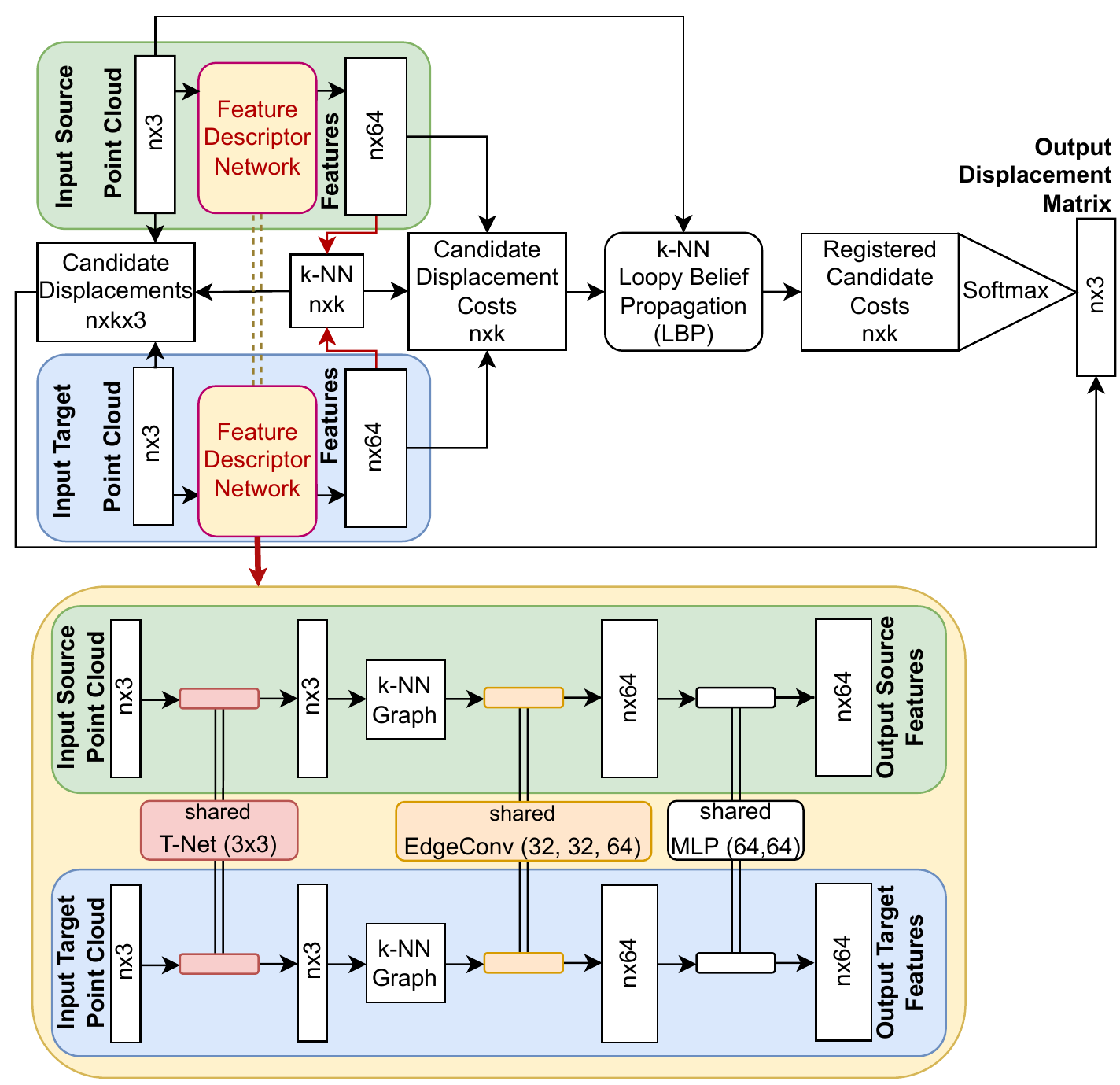}
\caption{The network architecture of the proposed method.} \label{fig:ProposedMethod}
\end{figure}

The non-rigid registration network, demonstrated in Figure \ref{fig:ProposedMethod}, is built upon a GCNN and utilizes LBP. Our proposed architecture takes inspiration from \cite{hansen2021deep} and incorporates a novel feature descriptor network to enhance robustness without compromising accuracy and computational efficiency. In \cite{hansen2021deep}, the method focuses on registering key points of deformed lungs. We use it as a foundation for our proposed method, Robust-DefReg, aiming to enhance its capabilities to serve as a general registration technique for various types of point clouds, not only medical data. Furthermore, our approach aims to make the registration process robust to rotation and other challenges simultaneously.\par

As demonstrated in Figure \ref{fig:ProposedMethod}, our novel feature descriptor is initially applied to the input point clouds, generating feature vectors of length 64 for each individual point. These feature vectors capture the local shape and structure of the surfaces represented within the point clouds. To achieve this, we integrate our network with T-Net \cite{jaderberg2015spatial}. Figure \ref{fig:ProposedMethod} illustrates the incorporation of T-Net to construct a new feature network. In the initial stage, a compact shared $3x3$ T-Net is introduced to align the coordinates of the two input point clouds. Subsequently, k-NN graphs are generated based on the transformed point clouds, and three EdgeConv layers are applied. These layers extract relevant features from the graphs. Finally, a fully connected linear MLP produces the output features. Further details on the utilized T-Net and EdgeConv architectures are provided in the supplementary paper. \par

Subsequently, after generating the feature vectors, a K-Nearest Neighbor (k-NN) search is conducted to identify the k nearest neighbors in the target point cloud for each point $\mathbf{x}_{i}$ in the source point cloud $X = \{\mathbf{x}_i \in \mathbb{R}^3 : i = 1, 2, ..., n\}$. The squared Euclidean distance between points serves as the metric for determining the closest neighbors. Once the neighbors are determined, they serve as candidates $\mathbf{c}^{p}_{i}$ for displacement, where $p \in \{1, 2, ..., k\}$, for each point $\mathbf{x}_{i}$ in the source point cloud. The displacements of these candidates represent potential transformations that can be applied to align the source point cloud with the target point cloud. The cost of displacement $d^{p}_{i}$ for each candidate $\mathbf{c}^{p}_{i}$ of point $\mathbf{x}_{i}$ is then computed based on the feature vectors. This cost measures the similarity between the features of the source point and the features of the target candidate. It can be expressed as:

\begin{equation}
d^{p}_{i} =\left\| f(\mathbf{x}_{i}) - f(\mathbf{c}^{p}_{i})\right\|_2^2
\label{eqn:d}
\end{equation}

Another k-NN graph is constructed in the source point cloud to represent the spatial relationships between the points and their neighbors. To address potential registration errors caused by noisy or missing correspondences, a robust regularization $r_{i j}^{p q}$ is introduced between neighboring points $\mathbf{x}_{i}$ and $\mathbf{x}_{j}$ in the graph and their candidates $\mathbf{c}^{p}_{i}$ and $\mathbf{c}^{q}_{j}$, where $q \in \{1, 2, ..., k\}$. This regularization penalizes deviations in relative displacements and is defined as:

\begin{equation}
r_{i j}^{p q}=\left\|\left(\mathbf{c}_i^p-\mathbf{x}_{i}\right)-\left(\mathbf{c}_j^q-\mathbf{x}_{j}\right)\right\|_2^2
\label{eqn:r}
\end{equation}

The Loopy Belief Propagation (LBP) algorithm is employed on the k-NN graph to refine the costs of the candidate displacements. LBP is a probabilistic inference algorithm that iteratively updates the belief of each point based on the beliefs of its neighbors by exchanging messages between them. The messages are updated iteratively, where the outgoing message $\mathbf{m}_{i \rightarrow j}^t$ from $\mathbf{x}_{i}$ to $\mathbf{x}_{j}$ at iteration $t$ is defined as:

\begin{equation}
\mathbf{m}_{i \rightarrow j}^t=\min _{1, \ldots, q, \ldots k}\left(\mathbf{d}_i+\alpha \mathbf{r}_{i j}^q-\mathbf{m}_{j \rightarrow i}^{t-1}+\sum_{(h, i) \in E} \mathbf{m}_{h \rightarrow i}^{t-1}\right)
\label{eqn:m}
\end{equation}

Here, $\alpha$ is a hyperparameter that controls the strength of the regularization $\mathbf{r}_{i j}^q=\left(r_{i j}^{1 q}, \ldots, r_{i j}^{p q}, \ldots, r_{i j}^{k q}\right)$, $\mathbf{m}_{i \rightarrow j}^0$ is initialized as 0, and $\mathbf{m}_{h \rightarrow i}^{t-1}$ represents the incoming messages from the neighbors $\mathbf{x}_h$ of the previous iteration. Finally, the Softmax function is applied to the registered candidate costs to normalize them into weights that reflect the confidence in the registration. These weighted candidate displacements are then combined to produce the final transformation for the source point cloud.\par

\section{Evaluation}
\begin{figure}[!ht]
\centering
\includegraphics[width=12.15cm, trim = 0.5cm 12.5cm 0cm 0cm, clip]{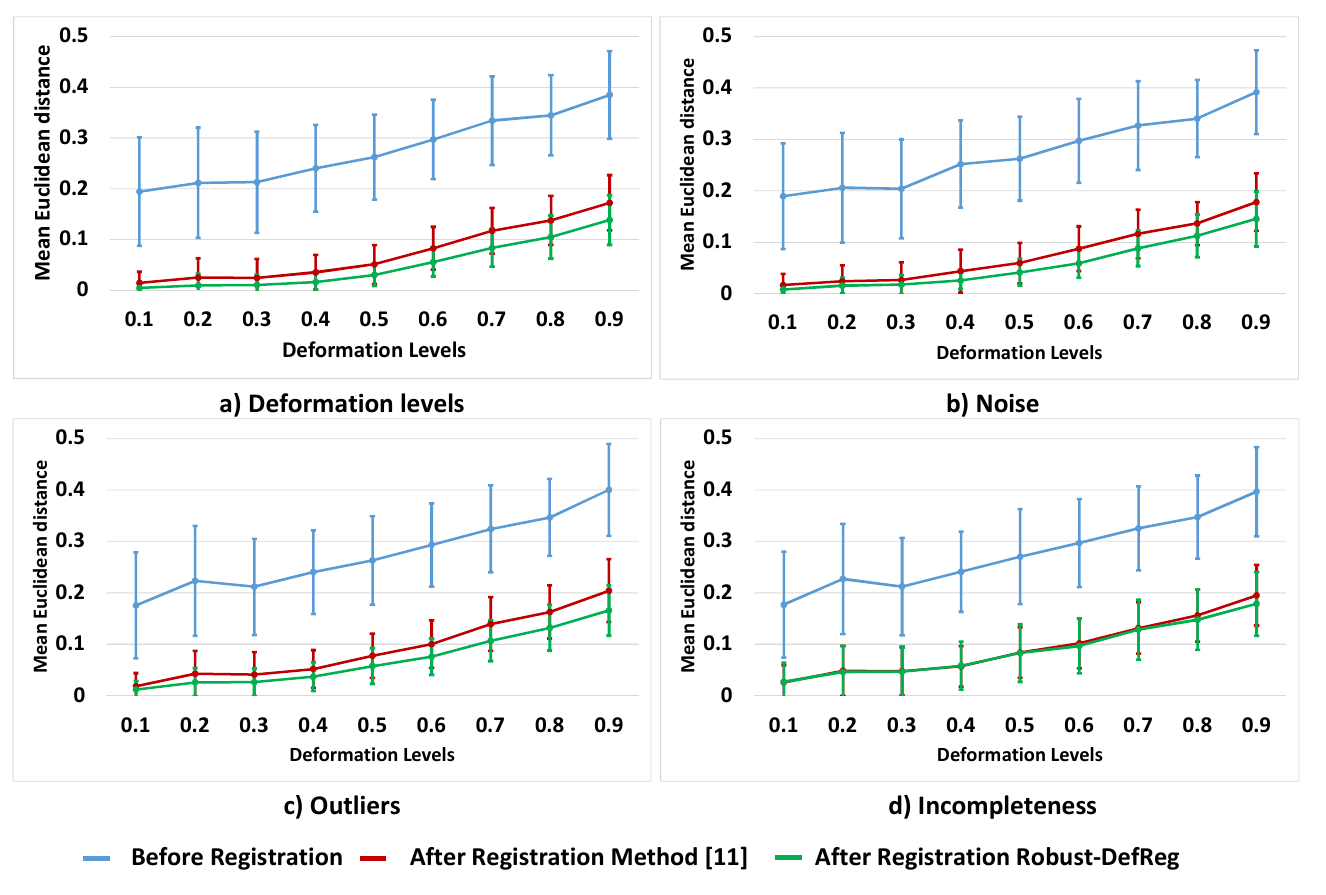}
\caption{Robustness to different deformation levels in comparison with method \cite{hansen2021deep}}
\label{fig:Evaluation1}
\end{figure}

An evaluation of the proposed method will be conducted in this section. Firstly, the computational efficiency of the method is presented. Next, the robustness of the proposed method is evaluated in terms of deformation levels, rotations, noise, outliers, and incompleteness. Two approaches are used to demonstrate the robustness of the proposed method compared to the state-of-the-art: 1) evaluating the method's robustness across different deformation levels and 2) selecting a specific deformation level and progressively increasing the challenge levels. The results are shown in Figures \ref{fig:Evaluation1} and \ref{fig:Evaluation2}.\par

These tests are conducted using the system with CPU (AMD Ryzen 7 7700X 8-Core @ 5.4GHz), GPU (NVIDIA GeForce RTX 4070Ti, 12 GB), RAM (Corsair DIMM 32 GB DDR5-6000), and SSD (Kingston KC3000 2048 GB). \par

The dataset utilized in this study for registration purposes is ModelNet10 \cite{wu20153d}. Each model in the dataset is sampled with 1024 points, normalized, and subjected to deformation. 

During training, the models are also randomly rotated up to 45 degrees around the z-axis. All these operations are automatically performed by the dataset loader at runtime. For training, $80\%$ of the dataset is randomly selected, while the remaining $20\%$ is reserved for validation. Due to the utilization of LBP in the final stage, a batch size of one is set, and all batch normalization layers are replaced with instance normalization. The LBP algorithm is executed for three iterations, with a regularization parameter $\alpha$ set to 50. The source k-NN is set to 10, and the target k-NN is set to 128. The networks are implemented using PyTorch \cite{paszke2017automatic} and are trained using the Adam optimizer with a learning rate of $e^{-4}$ and the GradScaler. The loss function employed is the L1 loss between corresponding points.\par

\subsection{Time Consumption}
Table \ref{tab:TotalTime} provides an overview of the training and registration times for each method. The training times for the networks are similar, with less than a $6\%$ difference between them. Similarly, the registration times show a variation of less than $5\%$.\par

\begin{table}[!htbp]
\caption{Computation times for the studied methods}
\label{tab:TotalTime}
\centering
\begin{tabular}{l l l}
\hline
Network & Training [h] & Registering [ms] \\
\hline
\hline
Method \cite{hansen2021deep} & 3:04 & 107.0 ± 1.7 \\
Robust-DefReg & 3:10 & 109.4 ± 1.2  \\
\hline
\end{tabular}
\end{table}

\subsection{Robustness to Different Deformation Levels}
To simulate deformation, we employ Thin Plate Spline (TPS) \cite{bookstein1989principal}. TPS is used to achieve a smooth deformation of a 3D surface by manipulating a set of control points. Detailed explanations of the deformed dataset creation are provided in the supplementary paper. \par

To evaluate the effectiveness of the proposed Robust-DefReg method, we compared its performance with the method introduced by \cite{hansen2021deep}. Figure \ref{fig:Evaluation1}(a) presents the Mean Euclidean distance between the registered source and target point clouds for both methods, as well as the distance between the initial source and target point clouds, plotted against deformation levels. 

The registration methods shown in the results demonstrate a reduction in distance compared to the distance before registration, and the shape of the curves generally follows the original distance curve. Moreover, the resulting distance achieved by Robust-DefReg is smaller than that of \cite{hansen2021deep}, and as the deformation level increases, the difference between the methods becomes more pronounced. When considering a distance of 0.1 as large, it can be observed that the method \cite{hansen2021deep} reaches this distance at a deformation level of 0.6, whereas our proposed method achieves it at a deformation level of 0.8. This indicates that our proposed method handles large deformations better than the state-of-the-art method. The comparison clearly highlights the significant improvement our proposed method offers over the base approach, effectively registering point clouds with a high level of accuracy.\par

\begin{figure}[!ht]
\centering
\includegraphics[width=12.15cm, trim = 1cm 13cm 0.5cm 0.5cm, clip]{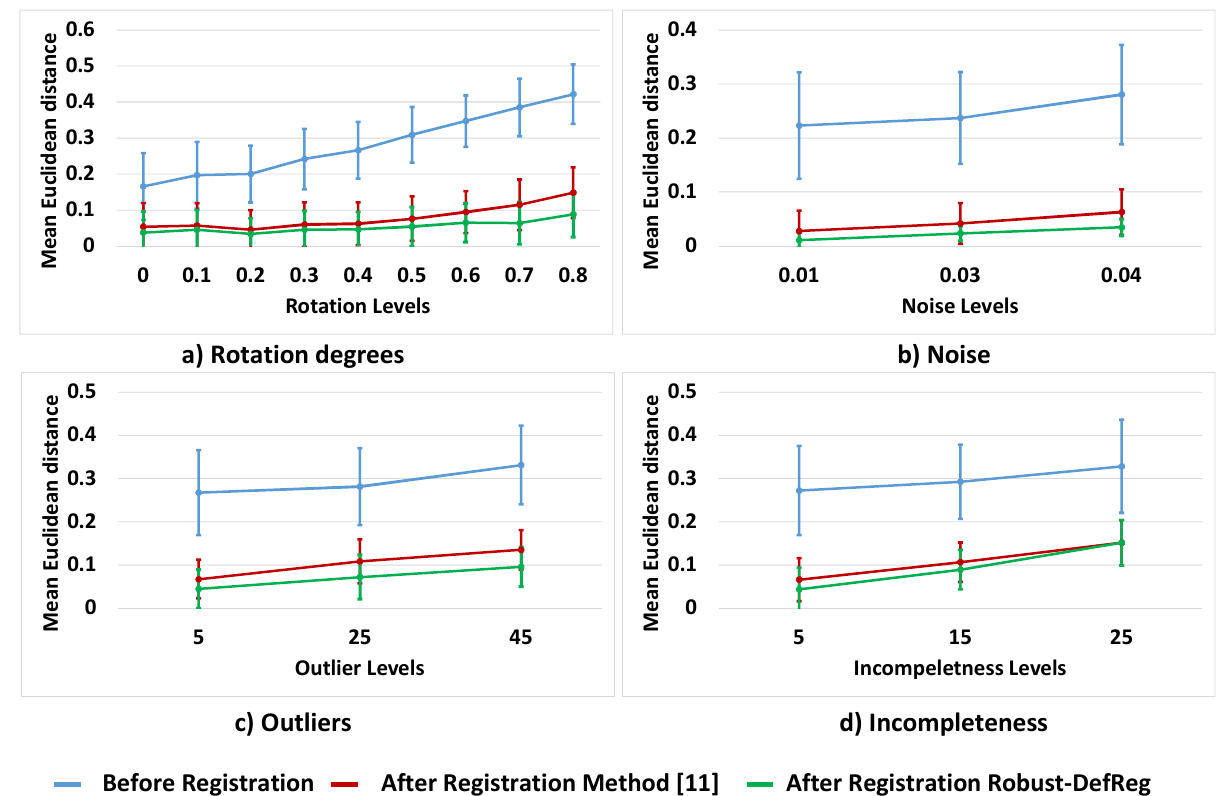}
\caption{Robustness to different levels of challenges in comparison with method \cite{hansen2021deep}}
\label{fig:Evaluation2}
\end{figure}

\subsection{Robustness to Rotation}

In this subsection, we evaluate the robustness of Robust-DefReg to rotation and compare it to the baseline approach \cite{hansen2021deep}. The results of this evaluation are shown in Figure \ref{fig:Evaluation2}(a). The rotation levels from 0 to 0.8 are defined in radians which randomly rotate objects around the z-axis.\par

It is worth noting that for all levels of rotation, the resulting distance of Robust-DefReg is consistently smaller than the distance obtained with \cite{hansen2021deep}. As shown in Figure \ref{fig:Evaluation2}(a), by increasing the rotation levels the distance results of the proposed method are always less than 0.1, however, the results shown for method \cite{hansen2021deep} are sharply increasing after rotation of more than 0.6 radians. \par

It can be concluded that after a certain degree of rotation, the corresponding points are no longer sufficiently close in 3D Euclidean space, requiring the registration process to rely on accurate feature descriptors to identify suitable candidates. The baseline approach fails to accomplish this effectively. In summary, the proposed method demonstrates its remarkable capability to significantly enhance the robustness to rotation, outperforming the baseline approaches in terms of maintaining accuracy even in challenging scenarios. \par

\subsection{Robustness to Noise}
Noise presents a significant problem in the registration of non-rigid point clouds, as it can have a negative impact on the accuracy and reliability of the registration outcomes. This noise can originate from various sources, such as limitations in sensors, errors in measurements, and environmental factors. It is essential to assess the robustness of the proposed method to different levels of noise in order to evaluate its performance in real-world scenarios. In this section, we examine the accuracy of the proposed method when exposed to varying noise levels. This evaluation aims to demonstrate the method's capability to handle the challenge of noise in conjunction with deformation and rotation. To accomplish this, random uniform displacements are added to the target point cloud, which has a shape of (n,3), up to a certain noise level. The resulting point cloud with noise, maintains the same dimensions as the original point cloud but includes perturbed versions of each point. More detail regarding generating the dataset with noise is explained in the supplementary paper. \par

Figure \ref{fig:Evaluation1}(b) presents the mean Euclidean distance as a function of deformation level with different noise levels added at each deformation level. Different noise levels and rotations up to 0.8 radians are averaged in the analysis. The observed approaches exhibit a relative resilience to noise, with only minor increases in error as the deformation levels rise. This phenomenon can be attributed to the fact that the noise perturbations have a minimal impact on the points, making them almost insignificant in the selection process of displacement candidates. Furthermore, both approaches employ Softmax in the subsequent LBP calculation, which contributes to the similarity in their results. However, still, the proposed method presents better accuracy at all deformation levels. Furthermore, when comparing Figure \ref{fig:Evaluation1}(a) and Figure \ref{fig:Evaluation1}(b), it can be observed that the approaches exhibit relative resilience to noise, as these graphs do not vary visibly. \par
In the second study demonstrated in Figure \ref{fig:Evaluation2}(b), three levels of noise are examined, and our proposed method consistently demonstrates higher accuracy compared to \cite{hansen2021deep}.\par

\subsection{Robustness to Outliers}
Apart from noise, outliers present a significant problem in non-rigid point cloud registration. These outliers refer to points that deviate substantially from the majority of the data and often arise from measurement errors or occlusions. Evaluating the robustness of the proposed method to outliers is crucial in assessing its effectiveness. In this section, we examine the accuracy of the proposed method when confronted with different ratios of outliers. This assessment aims to evaluate the method's ability to handle challenging scenarios involving deformation, rotation, and the presence of outliers. More information on how we generate outliers on the dataset is explained in the supplementary paper. 

The comparison between the proposed Robust-DefReg method and the baseline method \cite{hansen2021deep} is illustrated in Figure \ref{fig:Evaluation1}(c). The mean Euclidean distance is plotted against various deformation levels in a scenario where the average presence of outliers is considered. The analysis also includes an averaged rotation of up to 0.8 radians. The method \cite{hansen2021deep} achieves a distance of less than 0.1 up to a deformation level of 0.5, but beyond this threshold, its accuracy gradually decreases. On the other hand, the proposed method maintains a distance of less than 0.1 up to a deformation level of 0.7. Additionally, comparing \ref{fig:Evaluation1}(a) and \ref{fig:Evaluation1}(c), it can be observed that the approaches exhibit relative resilience to outliers, as these graphs do not vary visibly.\par

Furthermore, in another study shown in Figure \ref{fig:Evaluation2}(c), the evaluation is conducted based on increasing outlier levels. The results demonstrate that the proposed method achieves better accuracy compared to the state of the art as the number of outliers increases. This can be attributed to the challenge posed by a high percentage of outliers, which makes it difficult to identify correct correspondences within the feature space. Nevertheless, the proposed method consistently maintains higher accuracy in the presence of outliers across different deformation levels. \par

\subsection{Robustness to Incompleteness}
In addition to noise, outliers, and rotation, incompleteness is another significant factor that affects non-rigid point cloud registration. Incompleteness refers to missing or partially captured data in the point cloud, which can occur due to occlusions, sensor limitations, or other factors. Assessing the robustness of the proposed method to incompleteness is crucial for evaluating its effectiveness in real-world scenarios. In this section, we examine the accuracy of the proposed method when faced with varying levels of incompleteness. This evaluation aims to assess the method's ability to handle this challenge along with rotation. To simulate incompleteness, patches of points are randomly deleted from the source point clouds, which have a size of (n,3). Detailed information on generating the incompleteness dataset is provided in the supplementary paper.

Figure \ref{fig:Evaluation1}(d) displays a comparative analysis of Robust-DefReg and \cite{hansen2021deep} regarding their ability to handle incomplete point clouds. The plot shows the mean Euclidean distance for deformation levels up to 0.9 and rotations up to 0.8 radians, with the results being averaged. The observed decline in performance can be attributed to the absence of information from the missing points during the LBP calculation. This absence of data leads to imprecise correspondences in the feature space. Additionally, the proposed method's reliance on identifying candidates in the feature space amplifies the inaccuracies in the matches. Consequently, the results of the proposed method are not superior to those of the method \cite{hansen2021deep}. Figure \ref{fig:Evaluation2}(d) demonstrates the accuracy of the methods under different incompleteness levels, which show similar accuracy levels to the baseline method. \par

\section{Discussion}
In comparison to existing methods, Robust-DefReg offers several distinct advantages. Firstly, while coarse feature-based methods provide a rapid alignment, they may lack precision and struggle with capturing fine details in the point clouds. Conversely, fine feature-based methods focus on refining the initial alignment iteratively, resulting in more precise alignment but increased computational complexity. Robust-DefReg combines the benefits of both approaches by starting with a coarse alignment and then performing fine registration in further layers. This allows for accurate alignment of point clouds, including capturing fine details, without sacrificing computational efficiency.\par

While some existing methods demonstrate robustness to individual challenges, there is a lack of methods that simultaneously achieve high accuracy in all these scenarios. Robust-DefReg takes a step towards addressing this gap by presenting a method that is robust to different levels of deformation, noise, outliers, and data incompleteness. This robustness is achieved without significantly increasing the time complexity, making it a practical solution for real-world applications.\par

Furthermore, Robust-DefReg introduces an end-to-end method for non-rigid point cloud registration based on GCNN and LBP. This approach leverages the graph structure and neighboring nodes in a convolutional manner, allowing for the learning of geometric features from point clouds with neighborhood enabling accurate registration.\par

Finally, Robust-DefReg outperforms existing state-of-the-art methods in terms of accuracy under large deformation levels. By combining coarse and fine alignment strategies, Robust-DefReg achieves higher accuracy in non-rigid point cloud registration. \par

In summary, Robust-DefReg offers a unique combination of efficiency, accuracy, and robustness in non-rigid point cloud registration. By integrating coarse and fine alignment strategies, leveraging GCNNs and LBP, and addressing various challenges, Robust-DefReg represents a significant advancement in the field of point cloud registration.\par

\section{Conclusion}
In conclusion, this paper proposed a robust non-rigid point cloud registration method called Robust-DefReg, based on T-Net, Graph Convolutional Neural Networks (GCNN), and Loopy Belief Propagation (LBP). The method addresses the challenges of non-rigid registration by combining coarse and fine feature-based approaches within an end-to-end pipeline, enabling appropriate initial alignment and subsequent fine registration. Overall, Robust-DefReg presents accurate and robust alignment results. Future work may focus on expanding the method's applicability to more complex scenarios and exploring potential optimizations for further improving its performance.\par

\bibliographystyle{splncs04}
\bibliography{XXX-main.bbl}

\end{document}